\documentclass{llncs}
\usepackage{makeidx}  % allows for indexgeneration
\usepackage{amsmath}
   \usepackage{amsfonts}   % if you want the fonts
   \usepackage{amssymb}    % if you want extra symbols%
\title{APRIL: Active Preference-learning based Reinforcement Learning}
\titlerunning{APRIL: Active Preference-learning based Reinforcement Learning}  % abbreviated title (for running head)
\author{Riad Akrour \and Marc Schoenauer \and Mich\`ele Sebag}
\authorrunning{Riad Akrour et al.} % abbreviated author list (for running head)

\institute{TAO, CNRS $-$ INRIA $-$ LRI\\
Universit\'e Paris-Sud, F-91405 Orsay Cedex\\
\email{FirstName.Name@lri.fr}}
\def\EE{{\rm I\hspace{-0.50ex}E}}
\def\RR{{\rm I\hspace{-0.50ex}R}}
\usepackage{graphicx}
\def\APRIL{{\sc April}}
\def\PPL{{\sc Ppl}}
\def\Arch{\mbox{${\cal U}_t$}}
\def\x{\mbox{\bf x}}
\def\u{\mbox{\bf u}}
\def\w{\mbox{\bf w}}
\def\J{\mbox{${J}_t$}}

\begin{document}

\maketitle              % typeset the title of the contribution

\begin{abstract}
This paper focuses on reinforcement learning (RL) with limited prior knowledge. In the domain of swarm robotics for instance, the expert can hardly design a reward function or demonstrate the target behavior, forbidding the use of both standard RL and inverse reinforcement learning. Although with a limited expertise, the human expert is still often able to emit preferences and rank the agent demonstrations. Earlier work has presented an iterative preference-based RL framework: expert preferences
are exploited to learn an approximate policy return, thus enabling the agent to achieve direct policy search. Iteratively, the agent selects a new candidate policy and demonstrates it; the 
expert ranks the new demonstration comparatively to the previous best one; the expert's ranking feedback enables the agent to refine the approximate policy return, and the process is iterated. \\
In this paper, preference-based reinforcement learning is combined with active ranking in order to 
decrease the number of ranking queries to the expert needed to yield a satisfactory policy. 
Experiments on the mountain car and the cancer treatment testbeds witness that a couple of dozen rankings 
enable to learn a competent policy.
\keywords{reinforcement learning, preference learning, interactive optimization, robotics}
\end{abstract}

\section{Introduction}
\def\x{\mbox{\bf x}}
\label{sec:introduction}
Reinforcement learning (RL) \cite{Sutton_Barto_1998,Szepesvari} raises a main issue, that of the prior knowledge 
needed to efficiently converge toward a (nearly) optimal policy. Prior knowledge can 
be conveyed through the smart design of the state and action space, addressing the limited scalability of RL algorithms. The human expert can directly demonstrate some optimal or nearly-optimal behavior, 
speeding up the acquisition of an appropriate reward function and/or the exploration of the RL search space through inverse reinforcement learning \cite{Ng-Russell2000}, learning by imitation \cite{Billard}, or learning by demonstration \cite{Konidaris10}. The use of preference learning,
allegedly less demanding for the expert than inverse reinforcement learning, has also been investigated
in RL, respectively to learn a reward function \cite{Furnkranz11} or a policy return function \cite{PPL11}. 
In the latter approach, referred to as preference-based policy learning and motivated by swarm robotics, 
the expert is unable to design a reward function or demonstrate an appropriate behavior; the expert is 
more a knowledgeable person, only able to judge and rank the behaviors demonstrated by the learning agent. Like inverse reinforcement learning, preference-based policy learning learns a policy return; but demonstrations only rely on the learning agent, while the expert 
provides feedback by emitting preferences and ranking the demonstrated behaviors (section \ref{sec:soa}). 

Resuming the preference-based policy learning (\PPL) approach \cite{PPL11}, the contribution of the 
present paper is to extend \PPL\ along the lines of active learning, in order to minimize the number of expert's ranking feedbacks needed to learn a satisfactory policy. However our primary goal is 
to learn a competent policy; learning an accurate policy return is but a means to learn an accurate policy.
More than active learning {\em per se}, our goal thus relates to interactive optimization \cite{Brochu} and online recommendation \cite{Viappiani10}. The Bayesian settings used in these related works
(section \ref{sub:bayes}) will inspire the proposed {\em Active Preference-based 
Reinforcement Learning} (\APRIL) algorithm. 

The difficulty is twofold. Firstly, the above Bayesian approaches hardly scale up to large-dimensional
continuous spaces. Secondly, the \PPL\ setting requires one to consider two different search spaces. Basically, RL is a search problem on the policy space $\cal X$, mappings of the state space on the action space. 
%\footnote{Equivalently, policies can be represented via value functions, mapping the state $\times$ action space onto $\RR$ \cite{Sutton_Barto_1998}. Only deterministic RL settings will be considered in the following.}. 
However, the literature underlines that complex policies can hardly be expressed in the state $\times$ action space for tractability reasons \cite{PSR}. A thoroughly investigated  alternative is to use parametric representations (see e.g. \cite{PeterSchaal08} among many others), for instance using the weight vectors of a neural net as policy search space $\cal X$ (${\cal X} \subset \RR^d$, with $d$ in the order of thousands). 
Unfortunately earlier experiments suggest that parametric policy representations might be ill-suited to learn a preference-based policy return \cite{PPL11}. The failure to learn an accurate preference-based 
policy return on the parametric space is explained as the expert's preferences essentially relate
to the policy behavior, on the one hand, and the policy behavior depends in a highly non-smooth way on its parametric description on the other hand. Indeed, small modifications of a neural weight vector $\x$ can entail arbitrarily large differences in the behavior of policy $\pi_x$, depending on the robot environment (the tendency to turn right or left in front of an obstacle might have far fetched impact on the overall robot behavior). 

The \PPL\ framework thus requires one to simultaneously consider the parametric representation of 
policies (the primary search space) and the behavioral representation of policies (where the policy return, a.k.a. objective to be optimized, can be learned accurately). The distinction between the parametric and the behavioral spaces is reminiscent of the distinction between the input and the feature spaces, at the core of the celebrated kernel trick \cite{Cortes}. Contrasting with the kernel framework however, the mapping $\Phi$ (mapping the parametric representation $\x$ of policy $\pi_x$ onto the behavioral description $\Phi(\x)$ of policy $\pi_x$) is non-smooth\footnote{Interestingly, policy gradient methods face the same difficulties, and the guarantees they provide rely on the assumption of a smooth $\Phi$ mapping  \cite{PeterSchaal08}.}.
In order for \PPL\ to apply the abovementioned Bayesian approaches used in interactive optimization \cite{Brochu} or online recommendation \cite{Viappiani10} where the objective function is defined on the search space, one should thus solve the inverse parametric-to-behavioral mapping problem and compute $\Phi^{-1}$. However, computing such inverse mappings is notoriously difficult in general \cite{TsochantaridisJHA05}; it is even more so in the RL setting as it boils down to inverting the generative model. 

The technical contribution of the paper, at the core of the \APRIL\ algorithm, is to propose a tractable approximation of the Bayesian setting used in \cite{Brochu,Viappiani10}, consistent with the parametric-to-behavioral mapping. The robustness of the proposed approximate active ranking criterion is 
first assessed on an artificial problem. Its integration within \APRIL\ is thereafter studied and a proof of concept of \APRIL\ is given on the classical mountain car problem, and the cancer treatment testbed first introduced by \cite{CancerProblem}.

This paper is organized as follows. Section \ref{sec:soa} briefly presents \PPL\ for self-containedness and discusses work related to 
preference-based reinforcement learning and active preference learning. Section \ref{sec:april}
gives an overview of \APRIL. Section \ref{sec:resu} is devoted to the empirical validation of the approach and the paper concludes with some perspectives for further research.

\section{State of the art}
\label{sec:soa}
This section briefly introduces the notations used throughout the paper, assuming the reader's familiarity with reinforcement learning and referring to \cite{Sutton_Barto_1998} for a comprehensive presentation. Preference-based 
policy learning, first presented in \cite{PPL11}, is thereafter described for the sake of self-containedness, and discussed with respect to  inverse reinforcement learning \cite{Abbeel04,Kolter07} and preference-based value learning \cite{Furnkranz11}. Lastly, the section introduces related work 
in active ranking, specifically in interactive optimization
and online recommendation.

\subsection{Formal background}
Reinforcement learning classically considers a Markov decision process framework $({\cal S}, {\cal A}, p,  r, \gamma, q)$, where $\cal S$ and $\cal A$ respectively denote the state and the action spaces, 
$p$ is the transition model ($p(s,a,s')$ being the probability of being in state $s'$ after selecting 
action $a$ in state $s$),  $r: {\cal S} \mapsto \RR$ is a bounded reward function, $0 < \gamma < 1$ is a discount factor, and $q: {\cal S} \mapsto [0,1]$ is the initial state probability distribution. To each policy $\pi$ ($\pi(s,a)$ being the probability of selecting action $a$ in state $s$), 
is associated policy return $J(\pi)$, the expected discounted reward collected by $\pi$ over time: 
\[ J(\pi) = \EE_{\pi, p, s \sim q} \left [ \sum_{h=0}^\infty \gamma^h r(s_h) ~|~ s_0 = s \right ] \]
RL aims at finding optimal policy $\pi^*= \mbox{~arg max~} J(\pi)$. 
Most RL approaches, including the famed value and policy iteration algorithms, rely on the fact that a value function $V_\pi: {\cal S} \mapsto \RR$ can be defined from any policy $\pi$,  and that a policy ${\cal G}(V)$ can be greedily defined from any value function $V$:
\begin{eqnarray} 
%\begin{align}
    V_\pi(s)  = r(s) + \gamma \sum_{a} \pi(s,a)  p(s,a,s')  V_\pi(s') \label{eq:V} \smallskip\\
{\cal G}(V)(s)  = \mbox{~arg max~} \{ V(p(s,a)), a \in {\cal A} \} \label{eq:pi}
%\end{align}
\end{eqnarray}
Value and policy iteration algorithms, alternatively updating the value function and the policy (Eqs. (\ref{eq:V}) and (\ref{eq:pi})),  provide convergence guarantees toward the optimal policy provided that the state and action spaces are visited infinitely many times \cite{Sutton_Barto_1998}. Another RL approach, referred to as direct policy learning \cite{PeterSchaal08}, proceeds by directly optimizing some objective function a.k.a. policy return on the policy space. 

\subsection{Preference-based RL}\label{sub:P}\label{sub:pref}
Preference-based policy learning (\PPL) was designed to achieve RL when the reward function is unknown and generative model-based approaches are hardly applicable. 
%there is no generative model available
%(the transition function is unknown, the empirical evidence is severely restricted), and the reward %function is unknown.
 As mentioned, the motivating application is swarm robotics, where simulator-based approaches
are discarded for tractability and accuracy reasons, and the {\em individual} robot reward is not known since the target behavior is defined at the collective swarm level.

\PPL\ is an iterative 3-step process. During the demonstration step, the robot demonstrates a policy; during the ranking step, the expert ranks the new demonstration comparatively to the previous best demonstration; during the self-training step, the robot updates its model of the expert preferences, and
determines a new and hopefully better policy. {\em Demonstration} and {\em policy trajectory} or simply {\em trajectory}  will be used interchangeably in the following.

Let  $\Arch = \{\u_0, \ldots \u_{t};~ (\u_{i_1} \prec \u_{i_2}), i = 1 \ldots t \}$ the archive of all demonstrations seen by the expert and all ranking constraints defined from the expert's preference
 up to the $t$-th iteration. A utility function \J\ is defined on the space of trajectories as 
\[ \J(\u) = \langle \w_t,\u \rangle \]
where weight vector $\w_t$ is obtained by standard preference learning, solving quadratic constrained optimization problem \ref{eq:P} \cite{TsochantaridisJHA05,Joachims05}:
 \begin{equation}
\begin{array}{rl}
\tag{P} \hspace*{.2in} \mbox{Minimize} \hspace*{.2in} & F(\w)) = \frac{1}{2} ||\w||^2_2 + C \sum_{1 \le i \le t } \xi_{i_1,i_2}\bigskip\\
 s.t. \mbox{ for all ~} 1 \le i \le t \hspace*{.2in} & \langle \w, \u_{i_2} \rangle - \langle \w, \u_{i_1} \rangle \ge 1 - \xi_{i,j} \\
 & \xi_{i_1,i_2} \ge 0
\
\end{array}
\label{eq:P}
\end{equation}
Utility \J\ defines a policy return on the space of policies, naturally defined 
as the expectation of $\J(\u)$ over all trajectories generated from policy $\pi$ and still noted 
\J\ by abuse of notations:
\begin{equation}
  \J(\pi) = \EE_{u \sim \pi}[ \langle \w_t,\u \rangle]
\label{eq:J}
\end{equation}

In \cite{PPL11}, the next candidate policy $\pi_{t+1}$ is determined by heuristically optimizing a weighted sum of the current policy return \J, and the diversity w.r.t. archive \Arch.
A more principled active ranking criterion is at the core of the \APRIL\ algorithm (section \ref{sec:april}). 

\subsection{Discussion}
Let us discuss \PPL\ with respect to inverse reinforcement learning (IRL) \cite{Abbeel04,Kolter07}. 
IRL is provided with an informed, feature-based representation of the state space $\cal S$ 
(examples of such features $\phi_k(s)$ are the instant speed of the agent or whether it bumps in a pedestrian in state $s$). IRL exploits the expert's demonstration $\u^* = (s^*_0~s^*_1~s^*_2 \ldots s^*_h \ldots)$ to iteratively learn a linear reward function $r_t(s) = \langle \w_t, \Phi(s) \rangle$ on the feature space. Interestingly, reward function $r_t$ also defines a utility function $J_t$ on trajectories: letting 
$\u = (s_0 s_1 \ldots s_h\ldots)$ be a trajectory, 
\[ J_t(\u) =  \sum_{h=0}^\infty \gamma^h \langle \w_t, \Phi(s^*_h) \rangle = \langle \w_t, \sum_{h=0}^\infty \gamma^h  \Phi(s^*_h) \rangle = \langle \w_t, \mu(\u) \rangle \]
where the $k$-th coordinate of $\mu(\u)$ is given by $\sum_{h=0}^\infty \gamma^h \phi_k(s_h)$.
As in Eq. (\ref{eq:J}), a policy return function on the policy space can be derived by setting 
$J_t(\pi)$ to the expectation of $J_t(\u)$ over trajectories \u\ generated from $\pi$. 
  
IRL iteratively proceeds by computing optimal policy $\pi_t$ from reward function $r_t$ (using standard RL \cite{Abbeel04} or using Gibbs-sampling based exploration \cite{Kolter07}), and 
refining $r_t$ to enforce that $J_t(\pi_k) < J_t(u^*)$ for $k=1 \ldots t$. The process is iterated until reaching the desired approximation level.

In summary, the agent iteratively learns a policy return and a candidate policy in both IRL and \PPL. 
The difference is threefold. Firstly, IRL starts with an optimal trajectory $u^*$ provided by the human expert (which dominates all policies built by the agent by construction) whereas \PPL\ is iteratively provided with bits of information (this demonstration is/isn't better than the previous best demonstration) by the expert. Secondly, in each iteration IRL solves an RL problem using a generative model, whereas \PPL\ achieves direct policy learning. Thirdly, IRL is provided with an informed representation of the state space. 

Let us likewise discuss \PPL\ w.r.t. preference-based value learning \cite{Furnkranz11}. 
For each state $s$, each action $a$ is assessed in \cite{Furnkranz11} by 
executing the current policy until reaching a terminal state (rollout). On the basis of these rollouts, actions are ranked conditionally to $s$ (e.g. $a <_s a'$); the authors advocate that action ranking is more flexible and robust than a supervised learning based approach \cite{Lagoudakis},  discriminating the best actions in the current state from the other actions. The main difference with \PPL\ thus is that \cite{Furnkranz11} defines an order relation on the action space depending on the current state and the current policy, whereas \PPL\ defines an order relation on the policy space.

\subsection{Interactive optimization}
\label{sub:bayes}
During the \PPL\ self-training step, the agent must find a new policy, expectedly relevant 
w.r.t. the current objective function $J_t$, with the goal of finding as fast as possible a (quasi) optimal solution policy. This same goal, cast as an interactive optimization problem, has been tackled by \cite{Brochu} and \cite{Viappiani10} in a Bayesian setting.

In \cite{Brochu}, the motivating application is to help the user quickly find a suitable visual rendering 
in an image synthesis context. The search space ${\cal X}=\RR^D$ is made of the rendering parameter vectors.
The system displays a candidate solution, which is ranked by the user w.r.t. the previous ones. The ranking constraints are used to learn an objective function, represented as a Gaussian process using a binomial probit regression model. 
The goal is to provide as quickly as 
possible a good solution, as opposed to, the optimal one. Accordingly, the authors use the Expected Improvement over the current best solution as optimization criterion, and they return the best vector out of a finite sample of the search space. They further note that returning the optimal solution, e.g. using the Expected Global Improvement criterion \cite{EGO} with a branch-and-bound method, raises 
technical issues on high-dimensional search spaces.

In \cite{Viappiani10}, the context is that of online recommendation systems. The system iteratively provides the user with a choice query, that is a (finite) set of solutions $S$, of which the user selects the
one she prefers. The ranking constraints are used to learn a linear utility function $J$ on a low dimensional search space ${\cal X}=\RR^D$, with $J(\x) = \langle \w,\x \rangle$ and $\w$ a vector in $\RR^D$. Within the Bayesian setting, the uncertainty about the utility function is expressed through a belief $\theta$ defining a distribution over the space of utility functions. % ($\equiv \RR^D$).

Formally, the problem of (iterated) optimal choice queries is to simultaneously learn the user's utility function, and present the user with a set of good recommendations, such that she can select one with maximal expected utility. Viewed as a single-step (greedy) optimization problem, the goal thus boils down to finding a recommendation $\x$ with maximal expected utility $\EE_\theta[\langle \w, \x \rangle]$. 
In a global optimization perspective however \cite{Viappiani10}, the goal is to find a set of recommendations $S = \{\x_1, \ldots, \x_k\}$
with maximum expected {\em posterior utility}, defined as the expected gain in utility of the 
next decision. The expected utility of selection (EUS) is studied under several noise models, 
and the authors show that the greedy optimization of EUS provides good approximation guarantees of the 
optimal query. 

In \cite{Viappiani12}, the issue of the maximum expected value of information (EVOI) is tackled, and the authors consider the following criterion, where $\x^*$ is the current best solution: select $\x$ maximizing 
\begin{equation}
  EUS(\x) = \EE_{\theta, x > x^*}[\langle \w,\x \rangle] + \EE_{\theta, x <x^*}[\langle \w,
\x^*\rangle ] 
\label{eq:wvs}
\end{equation}
Eq. (\ref{eq:wvs}) thus measures the expected utility of \x, distinguishing the case where $\x$ actually 
improves on $\x^*$ (l.h.s) and the case where $\x^*$ remains the best solution (r.h.s). 
This criterion can be understood by reference to active learning and the so-called splitting index criterion \cite{Dasgupta05}. Within the realizable setting (the solution lies in the version space of all hypotheses consistent with all examples so far), an unlabeled instance $\x$ splits the version space into two subspaces: that of hypotheses labelling $\x$ as positive, and that of hypotheses labelling $\x$ as 
negative. The ideal case is when instance $\x$ splits the version space into two equal size subspaces; querying $\x$ label thus optimally prunes the version space. In the general case, the splitting index associated to $\x$ is the relative size of the smallest subspace: the larger, the better. 
In active ranking, any instance $\x$ likewise splits the version space into two subspaces: the challenger subspace of hypotheses ranking $\x$ higher than the current best instance $\x^*$, 
and its complementary subspace. 
In the Bayesian setting, considering an interactive optimization goal, the stress is put on the expected 
utility of \x\ on the challenger subspace, plus the expected utility of $\x^*$ on the complementary 
subspace.

\section{\APRIL\ Overview}
\label{sec:april}
\def\AA{\mbox{$\Pi$}}
\def\bev{{\em behavioral representation}}
\def\SM{SMS$(\pi)$} 

\def\At{{\mbox{$\cal A$$_t$}}}
\def\Jt{{\mbox{$J$$_t$}}}
\def\Jtp{{\mbox{$J^+$$_t$}}}
\def\Jtm{{\mbox{$J^-$$_t$}}}
\def\Ct{{\mbox{AEUS$_t$}}}
\def\AEUS{{\mbox{AEUS}}}
\def\pit{{\mbox{$\pi^*$$_t$}}}
\def\pre{{rank-based policy estimate}}
\def\esp{\hspace*{.2in}}
\def\pix{\mbox{$\pi_x$}}
\def\zx{\mbox{{\bf u}$_x$}}
\def\phix{\mbox{$\phi(\pi_x)$}}
\def\Phix{\mbox{$\Phi(\x)$}}

\def\w{\mbox{\bf w}}
\def\wp{\mbox{{\bf w}$^+$}}
\def\wn{\mbox{{\bf w}$^-$}}
\def\z{\mbox{\bf u}}
\def\zt{\mbox{{\bf u}$_t$}}
\def\zo{\mbox{{\bf u}$_0$}}
%\def\zst{\mbox{{\bf u}$_t^*$}}
%XXX J'enleve le *, on a suppos'e que l'archive 'etait ordonn'ee.
\def\zst{\mbox{{\bf u}$_t$}}
\def\VS{\mbox{{\bf W}$_t$}}
\def\VSp{\mbox{{\bf W}$_t^+$}}
\def\VSn{\mbox{{\bf W}$_t^-$}}

Like \PPL,  Active Preference-based Reinforcement Learning (\APRIL)  is an 
iterative algorithm alternating a demonstration and a self-training phase. 
% The demonstration
% phase proceeds as in PPL: the selected policy is demonstrated to the expert, who ranks it comparatively to the current best trajectory $\u_t$; the \APRIL\ archive \Arch, including all demonstrated policies and ranking constraints, is updated accordingly. 
The only difference between \PPL\ and \APRIL\ lies in the self-training phase.
% , which determines the next candidate policy to be demonstrated to the expert according to criterion \Ct\ (section \ref{sec:C}).
This section first discusses the parametric and behavioral policy representations. It thereafter presents 
an approximation of the expected utility of selection criterion (\AEUS) used to select the next candidate policy to be demonstrated to the expert, which overcomes the intractability of the EUS criterion (section \ref{sub:bayes}) with regard to these two representations.

\subsection{Parametric and behavioral policy spaces}\label{sec:Feature}

As mentioned, \APRIL\ considers two search spaces. The first one noted ${\cal X}$, referred to as input space or parametric space, is suitable to generate and run the policies. In the following ${\cal X} = \RR^d$; policy \pix\ is represented by e.g. the weight vector $\x$ of a neural net or the parameters of a control pattern generator (CPG) \cite{CPG}, mapping the current sensor values onto the actuator values. 
As mentioned, the parametric space is ill-suited to learn a preference-based policy return,
as the expert's preferences only depend on the agent behavior and  
the  agent behavior depends in an arbitrarily non-smooth way on the parametric policy 
representation. 
Another space, noted $\Phi({\cal X})$ and referred to as feature space or behavioral space, thus needs be
considered. Significant efforts have been made in RL to design a feature space suitable to capture the state-reward dependency (see e.g. \cite{HachiyaS10}); in IRL in particular, the feature space encapsulates an extensive prior knowledge \cite{Abbeel04}. In the considered swarm robotics framework however, comprehensive prior knowledge is not available, and the lack of generative model implies that massive data are not available either to construct an informed representation.
 
The proposed approach, inspired from \cite{ORegan01}, takes advantage of the fact that the agent is given for free the data stream made of its sensor and actuator values, generated along its
trajectories in the environment (possibly after unsupervised dimensionality reduction). %\cite{Riedmiller_EWRL12}). 
A frugal online clustering algorithm approach (e.g. $\varepsilon$-means
\cite{Duda}) is used to define sensori-motor clusters. To each such cluster, referred to as sensori-motor
state (sms), is associated a feature. 
It thus comes naturally to describe a trajectory by the
fraction of overall time it spends in every sensori-motor state\footnote{The use of the time fraction is 
chosen for simplicity; one might use instead the {\em discounted} cumulative time spent in every sms.}.
Letting $D$ denote the number of sms, each trajectory \zx\ generated from \pix\ thus is represented as a unit vector in $[0,1]^D$ ($||\zx||_1 = 1$). The behavioral representation associated to parametric policy \x, noted \Phix, finally is the distribution over $[0,1]^D$ of all trajectories \zx\ generated from policy $\pix$ (reflecting the actuator and sensor noise, and the presence and actions of other robots in the swarm).

Note that behavioral representation $\Phi({\cal X})$ does not require any domain knowledge. Moreover, it is consistent despite the fact that the agent gradually discovers its sensori-motor space; new sms are added along the learning process as new policies are considered, but the value of new sms is consistently set to 0 for earlier trajectories.

%\footnote{Further work is concerned with adaptively refining the clustering granularity, e.g. along the same lines as \cite{Oudeyer}.}. 

\subsection{Approximate expected utility of selection}\label{sec:C} 
Let $\Arch = \{\u_0, \ldots \u_{t-1}; (\u_{i_1} \prec \u_{i_2}), i = 1 \ldots t \}$ denote the archive of all demonstrations seen by the expert up the $t$-th iteration, and the ranking constraints defined from the expert preferences. With no loss of generality, the best demonstration in \Arch\ is noted $\u_{t-1}$. 

In \PPL\, the selection of the next policy to be demonstrated was based on the policy return 
$J_t(\pi_x) = \EE_{u \sim \pi_x} [\langle \w_t,\z\rangle]$, with $\w_t$ solution of the problem (\ref{eq:P}) (section \ref{sub:P}). By construction however, $\w_t$ is learned from the trajectories in the archive; it does not reward the discovery of new sensori-motor states (as they are 
associated a 0 weight by $\w_t$). Instead of considering the only max margin solution $\w_t$, the intuition is to consider the version space \VS\ of all $\w$ consistent\footnote{While we cannot assume a realizable setting, i.e. the expert's preferences are likely to be noisy as noted by \cite{Viappiani10}, the number of ranking constraints is always small relatively to the number $D$ of sensori-motor states. One can therefore assume that the version space defined from \Arch\ is not empty.} with the ranking constraints in the archive ${\cal U}_t$, along the same line as the expected utility of selection (EUS) criterion \cite{Viappiani10} (section \ref{sub:bayes}). 

The EUS criterion cannot however be applied as such, since policy return 
\J\ and the version space refer to the 
behavioral, trajectory space whereas the goal is to select an element on the parametric space; furthermore, both the behavioral and the parametric spaces are continuous and high-dimensional. 
An approximate expected utility of selection is thus defined on the behavioral and the parametric 
spaces, as follows. 
Let \zx\ denote a trajectory generated from policy $\pix$. 
The expected utility of selection of \zx\ can be defined as in \cite{Viappiani10}, as the 
expectation over the version space of the max between the utility of \zx\ and 
the utility of the previous best 
trajectory \zst: 
\[ EUS(\zx) = \EE_{w~in~W_t} [max (\langle \w,\zx\rangle, \langle \w,\zst\rangle)] \]
Specifically, trajectory \zx\ splits version space \VS\ into a challenger version space noted \VSp\ (including all \w\ with  $\langle \w,\zx\rangle > \langle \w,\zst\rangle)$), 
and its complementary subspace \VSn. The expected utility of selection of \zx\ thus becomes:
%XXX J'enleve \w en subscript, qui fait pas beau, trop grand.
\[ EUS(\zx) = \EE_{w~in~\VSp} [\langle \w,\zx\rangle] + \EE_{w~in~\VSn} [\langle \w,\zst\rangle] \]
The expected utility of selection of policy \pix\ is naturally defined as the expectation of EUS(\zx) over all trajectories \zx\ generated from policy \pix:
\begin{equation} 
\begin{array}{ll}
EUS(\pix) & = \EE_{u_x \sim \pix} [ EUS(\zx) ]\\
 &  = \EE_{u_x \sim \pix} \biggl[ \EE_{w~in~\VSp} [\langle \w,\zx\rangle] + \EE_{w~in~\VSn} [\langle \w,\zst\rangle] \biggr] \\
  \end{array}
\label{eq:eus}
\end{equation}

Taking the expectation over all weight vectors \w\ in $W^+$ or $W^-$ is clearly intractable as \w\ ranges in a high or medium-dimensional continuous space. Two approximations are therefore 
considered, defining the approximate expected utility of selection criterion (\AEUS). The first one consists of approximating the center of mass of a version space by the center of the largest ball in this version space, taking inspiration from the Bayes point machine \cite{HerbrichGC01}. The center of mass of 
\VSp\ (respectively \VSn) is replaced by \wp\ (resp. \wn) the solution of problem (\ref{eq:P}) where constraint $\zx > \zst$ (resp. $\zx < \zst$) is added to the set of constraints in archive \Arch.
As extensively discussed by \cite{HerbrichGC01}, the SVM solution provides a good approximation of 
the Bayes point machine solution provided the dimensionality of the space is ``not too high`` (more about this in section \ref{sec:artif}).

The second approximation takes care of the fact that the two version spaces \VSp\ and \VSn\
are unlikely of equal probability (said otherwise, the splitting index might be arbitrarily low). In order to approximate $EUS(\zx)$, one should thus further estimate the probability of \VSp\ and \VSn. 
Along the same line, the inverse of the objective value $F(\wp)$ maximized by \wp\ (section \ref{sub:pref}, problem \ref{eq:P}) is used to estimate the probability of \VSp: the higher the objective value, the smaller the margin and the probability of \VSp. Likewise, the inverse of the objective value $F(\wn)$ maximized by \wn\ is used to estimate the probability of \VSn.

Finally, the approximate expected utility of selection of a policy \pix\ is defined as:

\begin{equation}
  \Ct(\pix) = \EE_{u \sim \pix} \biggl[ \frac{1}{F(\wp)} \langle \wp,\zx\rangle + \frac{1}{F(\wn)} \langle \wn,\u_t\rangle] \biggr] 
\label{eq:ct}
\end{equation}

\subsection{Discussion}
The fact that \APRIL\ considers two policy representations, the parametric and the behavioral 
or feature space, aims at addressing the expressiveness/tractability dilemma. On the one hand, 
a high dimensional continuous search space is required to express competent policies. But such 
high-dimensional search space makes it difficult to learn a preference-based policy return from a moderate number of rankings, keeping the expert's burden within reasonable limits. 
On the other hand, the behavioral space does enable to learn a preference-based policy return 
from the little available evidence in the archive
(note that the dimension of the behavioral space is controlled by \APRIL) although the behavioral 
description might be insufficient to describe a flexible policy. 

The price to pay for dealing with both search spaces lies in the two approximations needed to transform 
the expected utility of selection (Eq. (\ref{eq:eus})) into a tractable criterion (Eq. (\ref{eq:ct})), 
replacing the two centers of mass of version spaces \VSp\ and \VSn\ (i.e. the solutions of the Bayes point machine  \cite{HerbrichGC01}) with the solutions of the associated support vector machine problems, and estimating the probability of these version spaces from the objective values of the associated SVM problems.

\section{Experimental results}
\label{sec:goal}
This section presents the experimental setting followed to validate \APRIL.
Firstly, the performance of the approximate expected utility of selection (\AEUS) criterion 
is assessed in an artificial setting. Secondly, the performance of \APRIL\ is  assessed comparatively 
to inverse reinforcement learning \cite{Abbeel04} on two RL benchmark problems.

\subsection{Validation of the approximate expected utility of selection}\label{sec:aeus}\label{sec:artif}
\begin{figure*}[t!]
\begin{center}
\begin{tabular}{cc}
\includegraphics[width=0.5\textwidth]{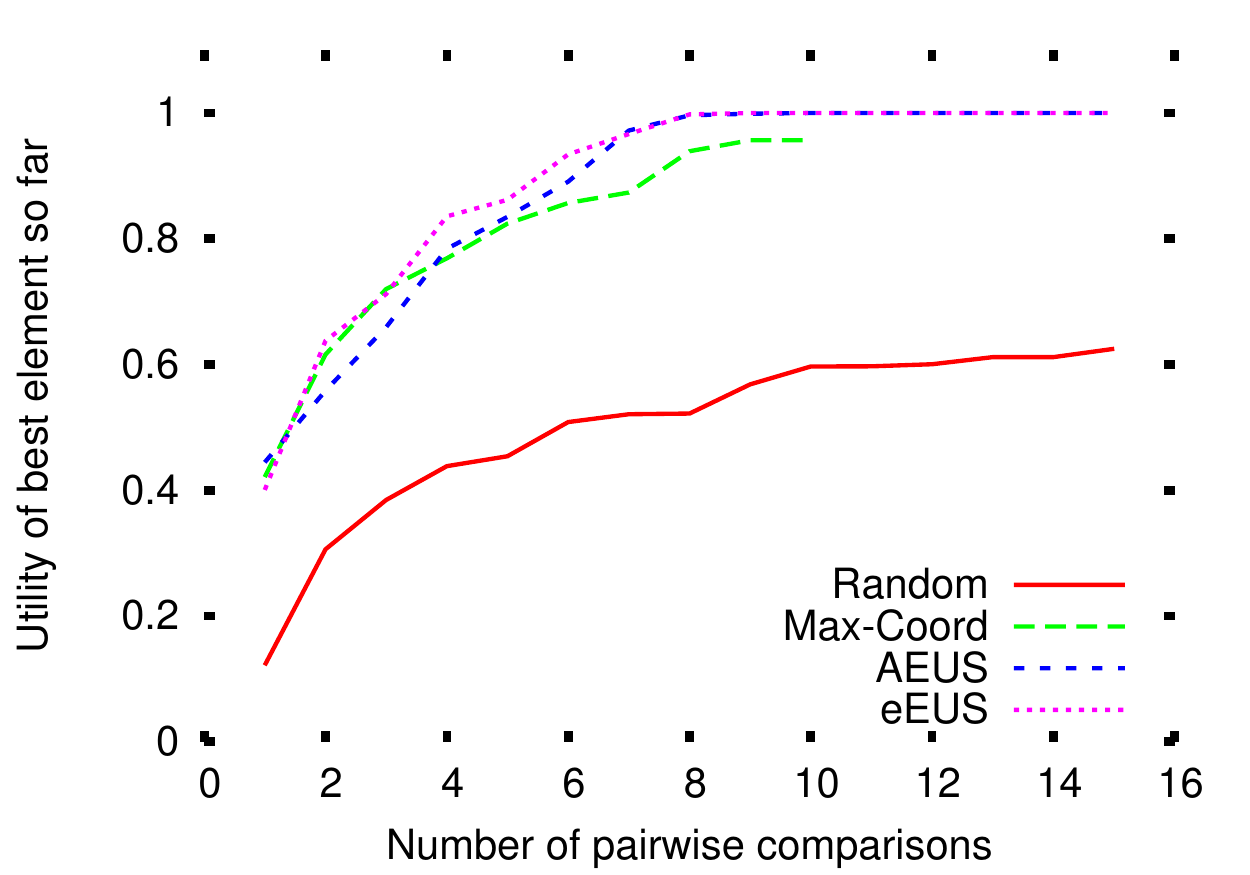}
&
\includegraphics[width=0.5\textwidth]{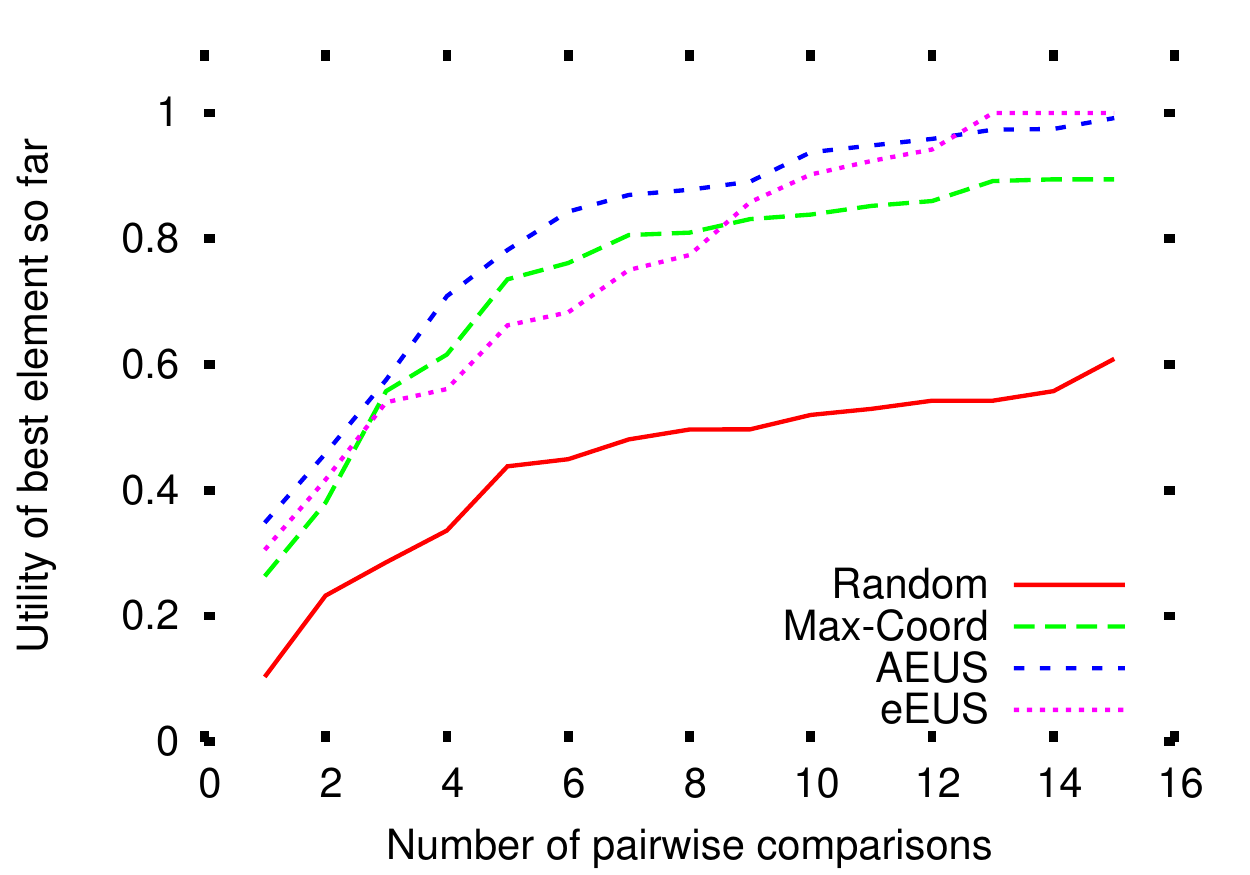}\\
$d=10$ & $d=20$\\
\includegraphics[width=0.5\textwidth]{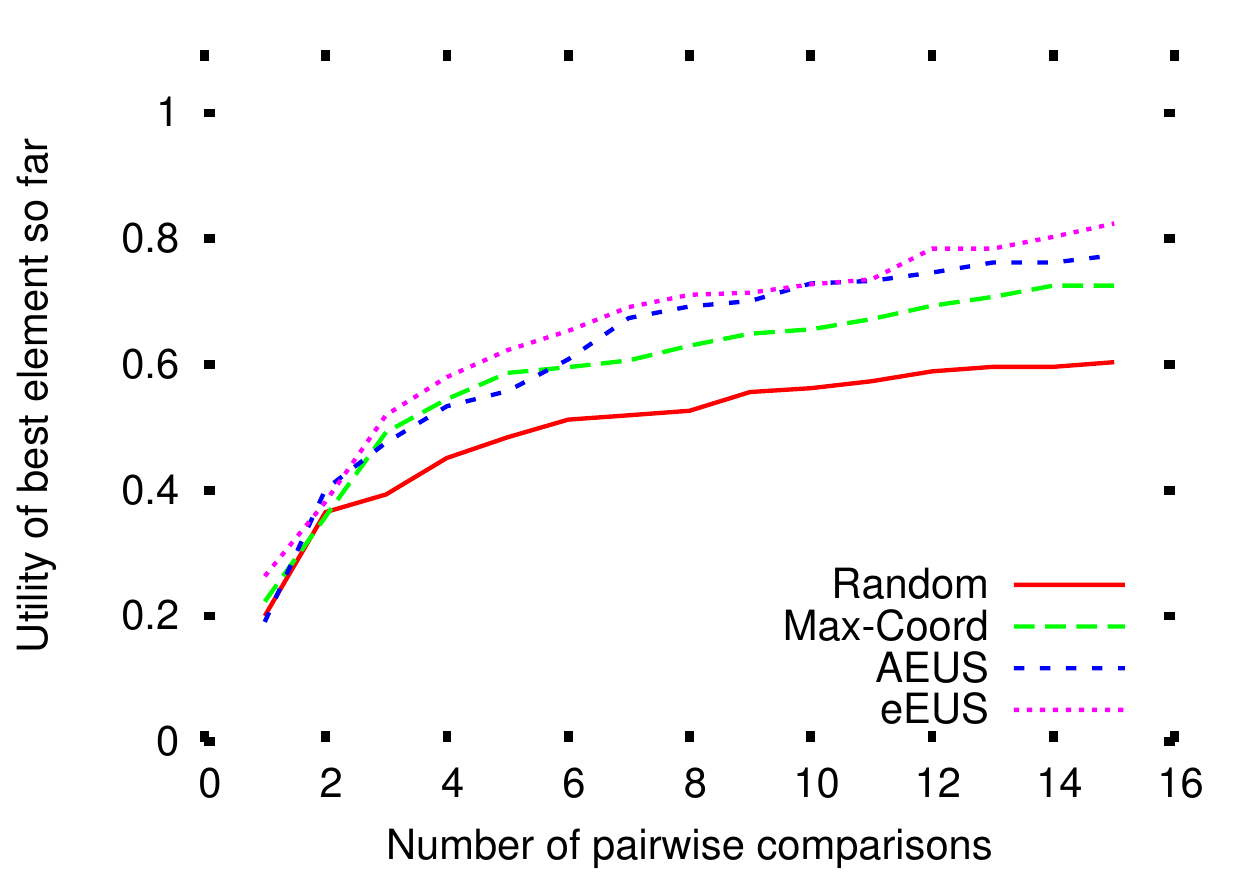}
&
\includegraphics[width=0.5\textwidth]{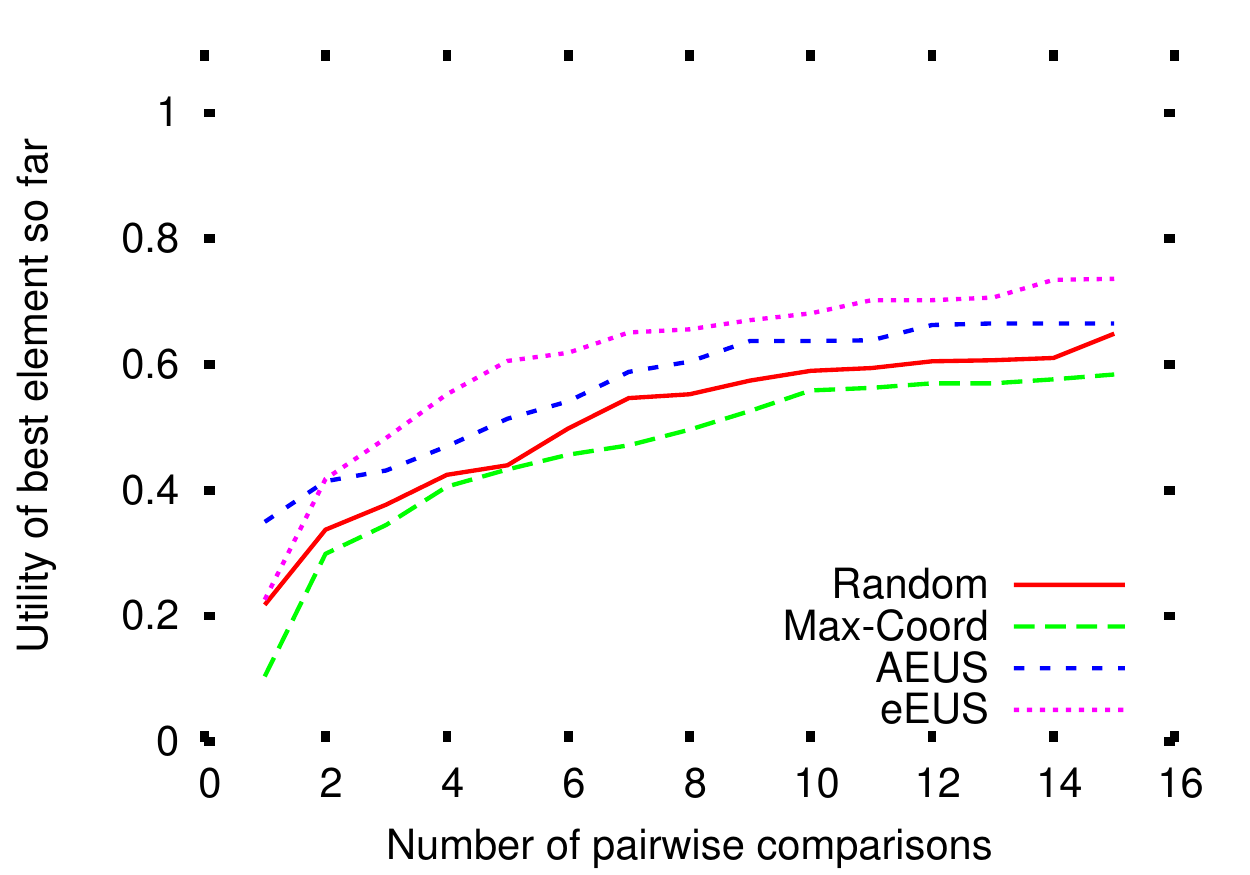}\\
$d=50$ & $d=100$\\
\end{tabular}
\end{center}
\caption{Performance of AEUS comparatively to baselines (see text)  {\em vs} number of 
pairwise comparisons,  depending on the dimension $d$ of the search space (results averaged over 101 runs).}

\label{fig_simplex}
\end{figure*}
The artificial active ranking problem used to investigate \AEUS\ robustness is inspired from the 
active learning frame studied by \cite{Dasgupta05}, varying the dimension $d$ of the space in 10, 20, 50, 100 (Fig. \ref{fig_simplex}).

In each run, a target utility function is set as a vector \w$^*$ uniformly selected in the $d$-dimensional $L_2$ unit sphere. A fixed sample $S = \{\u_1, \ldots \u_{1000} \}$ of 1,000 points uniformly generated\footnote{The samples are uniformly selected in the $d$-dimensional $L_1$  unit sphere, to account for the fact that the behavioral representation of a trajectory has $L_1$ norm 1 by construction (section \ref{sec:Feature}). } in the $d$-dimensional $L_1$  unit sphere is built.
At iteration $t$, the sample \u\ with best \AEUS\ is selected in $S$; the expert ranks it comparatively to the previous best solution \u$_{t}$, yielding a new ranking constraint (e.g. $\u < \u_{t}$), 
and the process is iterated. The \AEUS\ performance at iteration $t$ is computed as the scalar product of \u$_t$ and \w$^*$. 

\AEUS\ is compared to an empirical estimate of the expected utility of selection (baseline eEUS). In eEUS, the current best sample \u\ in $S$ is selected from Eq. (\ref{eq:eus}), computed from 10,000 points $\w$ selected in the version spaces \VSp\ and \VSn\ and $\u_t$ is built as above.
%\footnote{Note that \cite{Viappiani12} involves an additional optimization step, replacing \u$_t$ with the true optimum of EUS, whereas $\u_t$ is kept as is in the reported experiments.}.
Another two baselines are considered: Random, where sample \u\ is uniformly selected in $S$, and Max-Coord, selecting with no replacement \u\ in $S$ with maximal $L_\infty$ norm. The Max-Coord baseline was found to perform surprisingly well in the early active ranking stages, especially for small dimensions $d$. 
% 
% the current best 
% 
% . Iteratively, In each step ($x$ coordinate in Fig. \ref{fig_simplex}), the active
% ranking candidate \u$_t$ is selected out of a fixed set $S$ of 1,000 points uniformly sampled in the $d$-dimensional $L_1$  unit sphere (sampling in the $L_1$ sphere instead of the $L_2$ sphere for the point $\u \in S$ is justified by the fact that on policy learning problems, the behavioral representation of a policy also lays in the $L_1$  unit sphere as discussed in section \ref{sec:Feature}); the quality of \u$_t$ is estimated from its scalar product with \w$^*$; lastly, the expert provides a single bit of information, stating whether \u$_t$ improves on the current best solution. 
% The criteria used to select the active ranking candidate \u$_t$ out of $S$ are as follows. 
% The first one is criterion $C_t$ (Eq. \ref{eq:ct}). The second one drops the SVM approximations of $C_t$ for calculating the expectation over $W$ and instead sample uniformly 10,000 points $\w$ in the version spaces $W^+$ and $W^-$ to empirically estimate expectation $\EE_{\w~in~W^+}[\langle \w, \u_t \rangle]$ and $\EE_{\w~in~W^-}[\langle \w, \u_t^* \rangle]$. It is thus very close to the greedy optimization of $EUS$ in \cite{Viappiani12} (with the only difference being that in \cite{Viappiani12} $\u_t^*$ is not necessarily the best demonstrated point but the one having maximal expected utility requiring one additional optimization phase). 
% Two baseline approaches are also considered. The random baseline uniformly selects 
% \u\ in $S$ in each time step. The second baseline referred to as 
The empirical results reported in Fig. \ref{fig_simplex} show that \AEUS\ is a good active ranking criterion, yielding a good approximation of EUS. The approximation degrades gracefully as dimension $d$ increases: 
as noted by \cite{HerbrichGC01}, the center of the largest ball in a convex set yields a lesser good approximation of the center of mass thereof as dimension $d$ increases. In the meanwhile, the 
approximation of the center of mass degrades too as a fixed number of 10,000 points are used to 
estimate EUS regardless of dimension $d$. 
Random selection catches up as $d$ increases; quite the contrary, Max-Coord becomes worse as $d$ increases. 

\subsection{Validation of \APRIL}\label{sec:eAPRIL}\label{sec:resu}

The main goal of the experiments is to comparatively assess \APRIL\ with respect to inverse reinforcement learning (IRL \cite{Abbeel04}, section \ref{sec:soa}). Both IRL and \APRIL\ 
extract the sought policy through an iterative two-step processes. The difference is that IRL 
is initially provided with an expert trajectory, whereas \APRIL\ receives one bit of information 
from the expert on each trajectory it demonstrates (its ranking w.r.t. the previous best trajectory). \APRIL\ performance is thus measured in terms of 
``expert sample complexity'', i.e. the number of bits of information needed to catch up compared to IRL.
\APRIL\ is also assessed and compared to the black-box CMA-ES optimization algorithm, used with default parameters \cite{CMA-ES}.

All three IRL, \APRIL\ and CMA-ES algorithms are empirically evaluated on two RL benchmark problems, the well-known mountain car problem and the cancer treatment problem first introduced by \cite{CancerProblem}. None of these problems has a reward function.

Policies are implemented as 1-hidden layer neural nets with 2 input nodes and 1 output node, respectively
the acceleration for the mountain car (resp. the dosage for the cancer treatment problem). The hidden layer contains 9 neurons for the mountain car (respectively 99 nodes for the cancer problem), thus 
the dimension of the parametric search space is ${37}$ (resp. $397$). 

RankSVM is used as learning-to-rank algorithm \cite{Joachims_Implementation}, 
with linear kernel and $C = 100$.
All reported results are averaged out of 101 independent runs. 

\subsubsection{The cancer treatment problem.}
In the cancer treatment problem, a stochastic transition function is provided, yielding the next patient state from its current state (tumor size $s_t$ and toxicity level $t_t$) and the selected action (drug dosage level $a_t$):
\[ \begin{array}{ll}
s_{t+1} &= s_t + 0.15 \max(t_t, t_0)-1.2(a_t - 0.5) \times 1(s_t > 0) + \varepsilon\\
t_{t+1} &= t_t + 0.1 \max(s_t, s_0) + 1.2 (a_t - 0.5)   + \varepsilon  
   \end{array} \]
Further, the transition model involves a stochastic death mechanism (modelling the possible patient death by means of a hazard rate model).
The same setting as in \cite{Furnkranz11} is considered  with
three differences. Firstly, we considered a continuous action space (the dosage level is a real value in $[0,1]$), whereas the action space contains 4 discrete actions in \cite{Furnkranz11}. Secondly the time horizon is set to 12 instead of 6. Thirdly, a Gaussian noise $\epsilon$ with mean 0 and standard deviation $\sigma$ (ranging in 0, 0.05, 0.1, 0.2) is introduced in the transition model. The \AEUS\ of 
the candidate policies is computed as their empirical \AEUS\ average over 11 trajectories (Eq. \ref{eq:ct}).

The initial state is set to 1.3 tumor size and 0 toxicity. 
For the sake of reproducibility the expert preferences are emulated by favoring the trajectory with minimal sum of the tumor size and toxicity level at the end of the 12-months treatment. 

% \begin{figure}[b!]
% \centerline{\includegraphics[width=0.45\textwidth]{figures/cancer.pdf}}
% \caption{Comparative performances of \APRIL\, IRL and stochastic optimization on the cancer treatment problem, plotting the cumulative toxicity level and tumor size after 12-months treatment, versus the number of trajectories demonstrated policies to the expert. Results are averaged over 101 runs}
% \label{fig_cancer}
% \end{center}
% \end{figure}

\begin{figure*}[t!]
\begin{center}
\begin{tabular}{cc}
\includegraphics[width=0.5\textwidth]{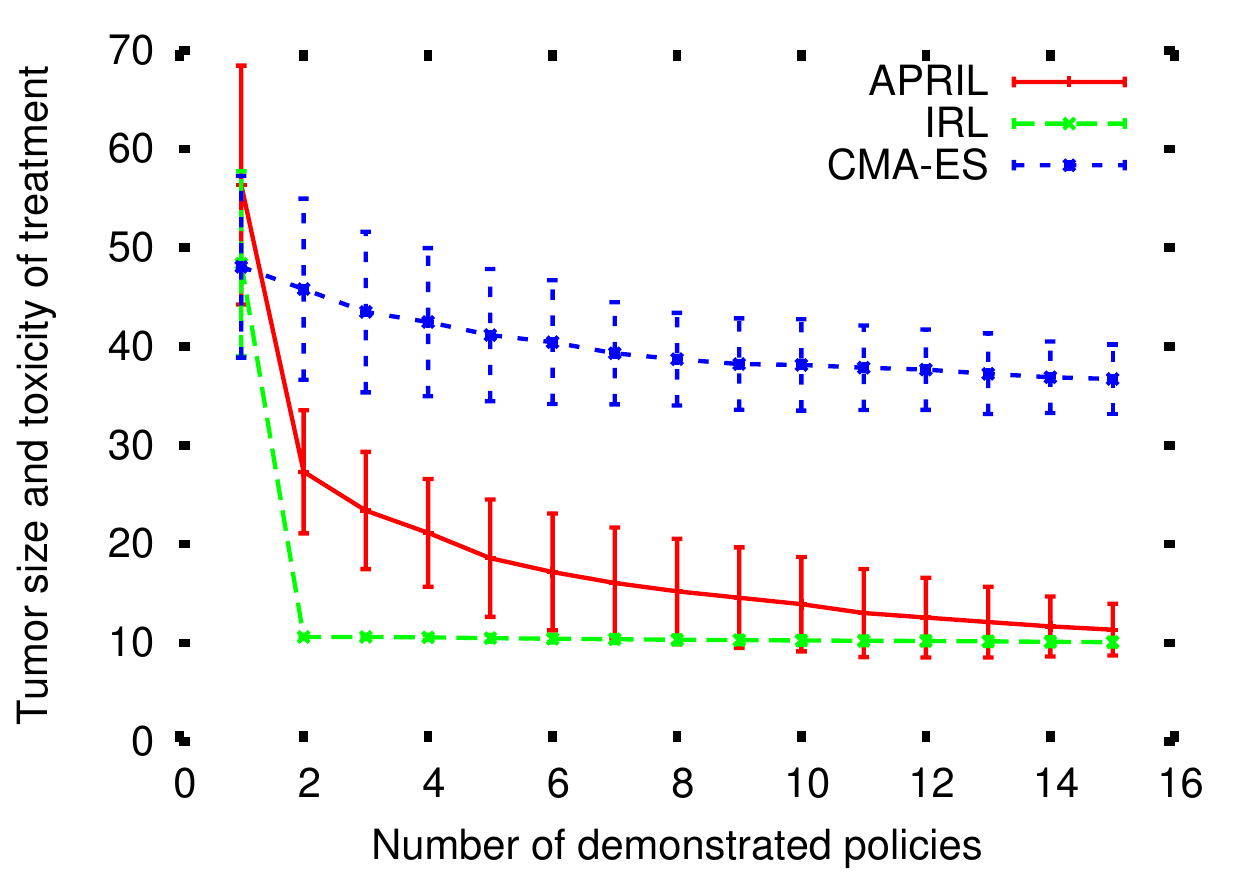}
&
\includegraphics[width=0.5\textwidth]{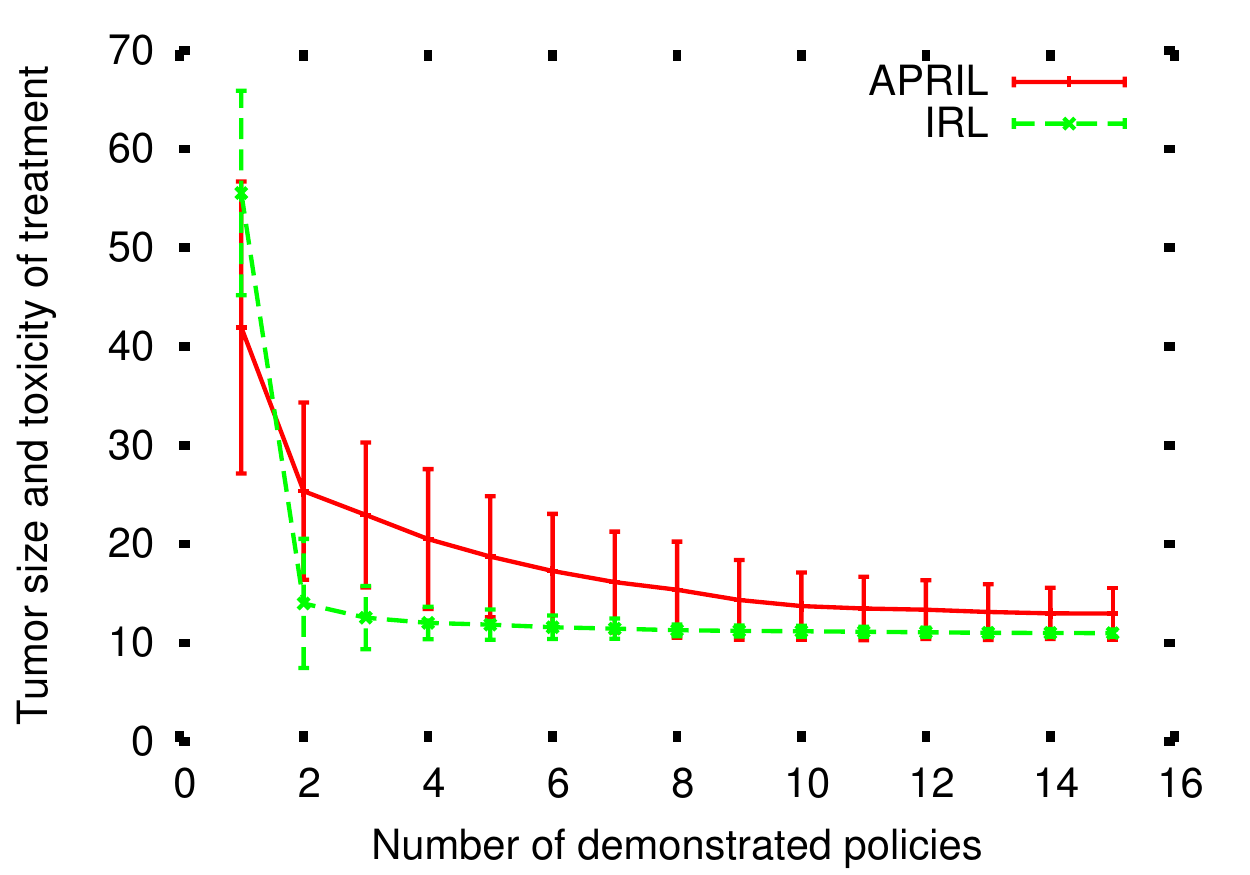}\\
$\sigma_{noise}=0$ & $\sigma_{noise}=0.05$\\
\includegraphics[width=0.5\textwidth]{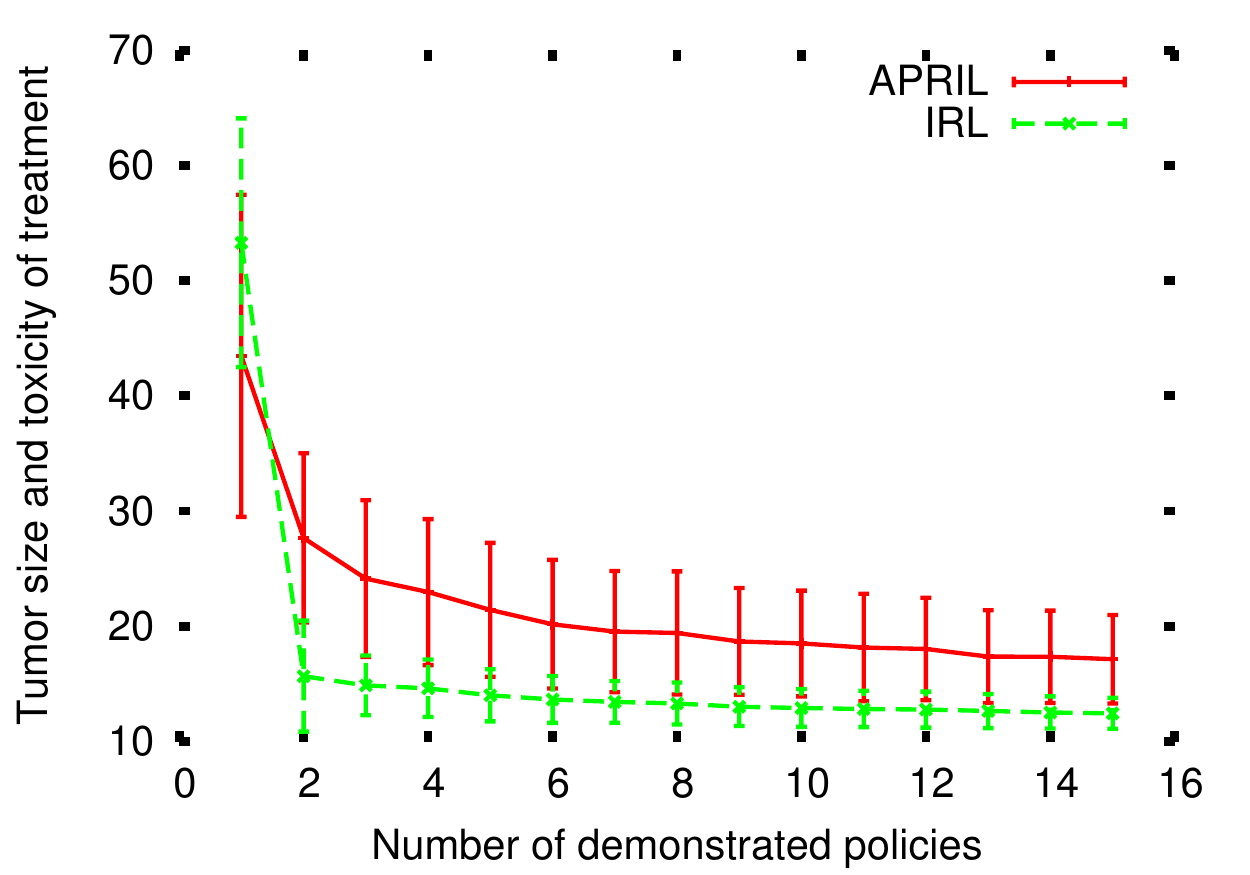}
&
\includegraphics[width=0.5\textwidth]{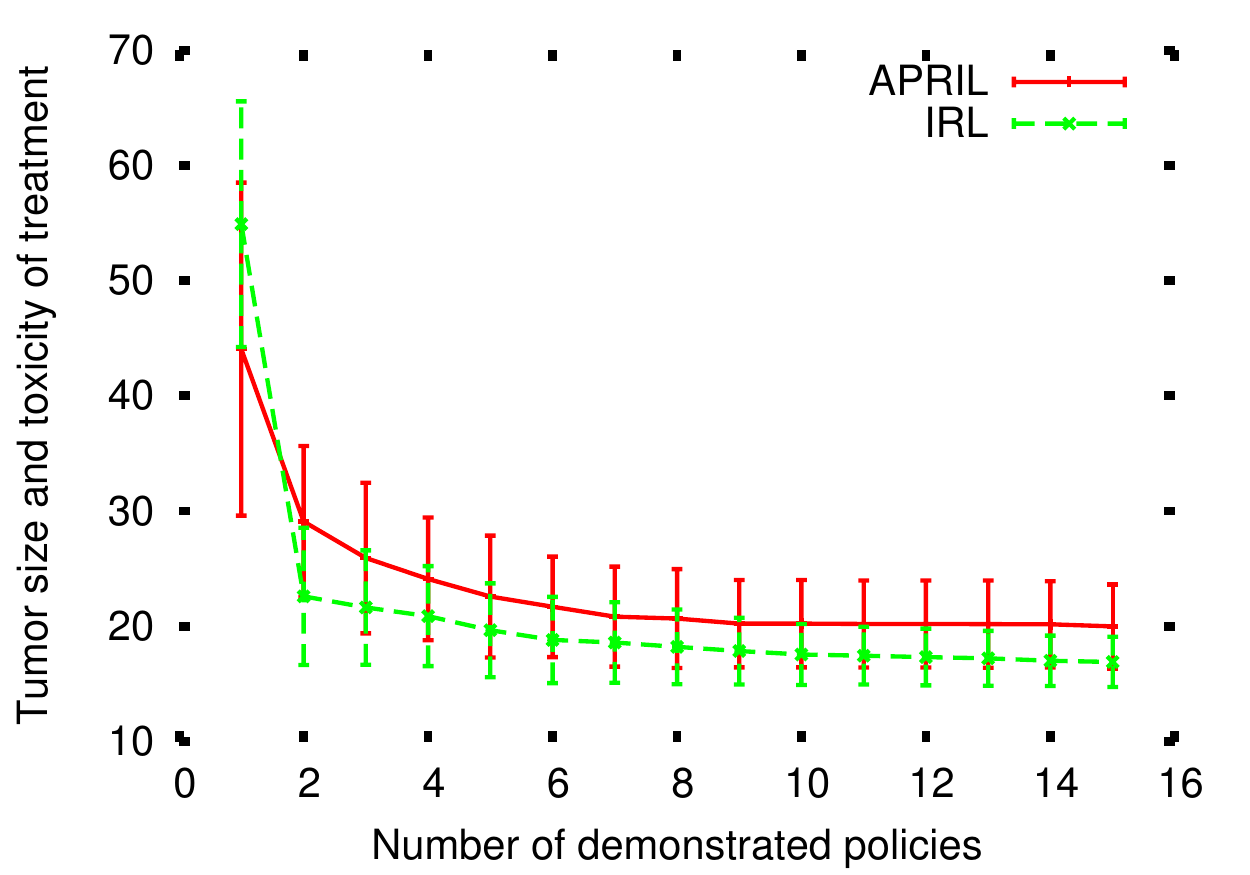}\\
$\sigma_{noise}=0.1$ & $\sigma_{noise}=0.2$\\
\end{tabular}
\end{center}
\caption{The cancer treatment problem: Average performance (cumulative toxicity level and tumor size after 12-months treatment) of \APRIL, IRL and CMA-ES versus the number of trajectories demonstrated to the expert, for noise level 0, .05, .1 and .2. Results are averaged over 101 runs.}
\label{fig_cancer}
\end{figure*}

The average performance (sum of tumor size and toxicity level) of the best policy found in each iteration is reported in Fig. \ref{fig_cancer}. It turns out that the cancer treatment problem is an easy problem for IRL, that finds the optimal policy in the second iteration. 
A tentative interpretation for this fact is that the target behavior extensively visits the 
state with zero toxicity and zero tumor size; the learned \w\ thus associates a maximal weight to this 
state. In subsequent iterations, IRL thus favors policies reaching this state as soon as possible. 
\APRIL\ catches up after 15 iterations, whereas CMA-ES remains consistently far 
from reaching the target policy in the considered number of iterations when there is no noise, and 
yields bad results (not visible on the plot) for higher noise levels.

\begin{figure}
\centerline{\includegraphics[width=0.55\textwidth]{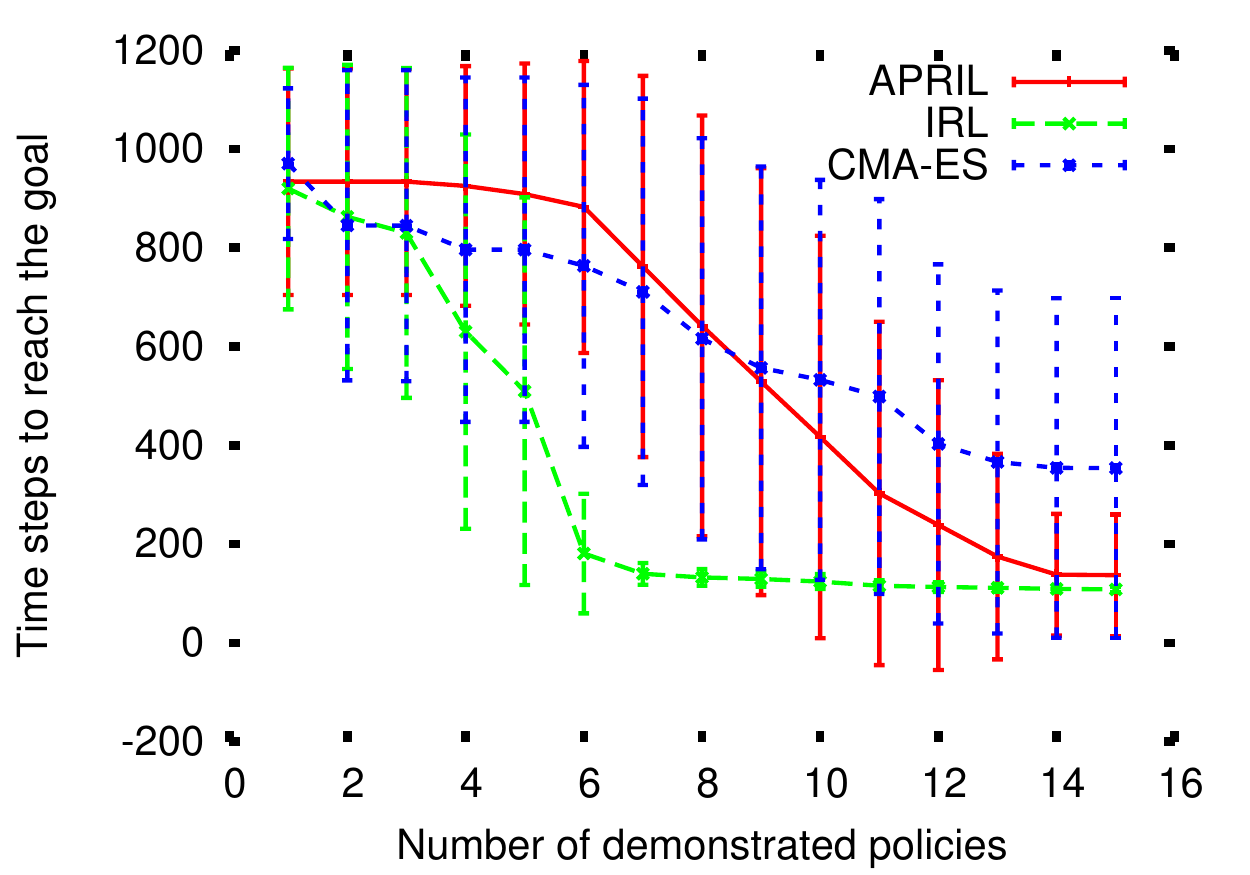}}
\caption{The mountain car problem: Average performance (number of time steps needed to reach the top of the mountain) of \APRIL, IRL and CMA-ES versus the number of trajectories demonstrated to the expert. Results are averaged over 101 runs.}
\label{fig_mc}
\end{figure}

\subsubsection{The mountain car.}

The same setting as in \cite{StoneRLEVO} is considered. The car state is described from its position and speed, initially set to position $-0.5$ with speed $0$. The action space is set to $\{-1, 0, 1\}$, setting the car acceleration. 
For the sake of reproducibility the expert preferences are emulated by favoring the trajectory which soonest reaches the top of the mountain, or is closest to the top of the mountain at some point 
during the 1000 time-step trajectory. 

Interestingly, the mountain car problem appears to be more difficult for IRL than
the cancer treatment problem (Fig.\ref{fig_mc}), which is blamed on the lack of expert features.
As the trajectory is stopped when reaching the top of the mountain and this state does not appear in the trajectory description, the target reward would have negative weights on every (other) sms feature. 
IRL thus finds an optimal policy after 7 iterations on average. 
As for the cancer treatment problem, \APRIL\ catches up after 15 iterations, while the stochastic 
optimization never catches up in the considered number of iterations. 

\section{Discussion and Perspectives}
The Active Preference-based Reinforcement Learning algorithm presented in this paper combines
Preference-based Policy Learning \cite{PPL11} with an active ranking mechanism aimed at decreasing the number of comparison requests to the expert, needed to yield a satisfactory policy. 

The lesson learned from the experimental validation of \APRIL\ is that a very limited external information might be sufficient to enable reinforcement learning: while mainstream RL requires a numerical reward to be associated to each state, while inverse reinforcement learning \cite{Abbeel04,Kolter07} requires the expert to demonstrate a sufficiently good policy, \APRIL\ requires a couple dozen bits of information (this trajectory improves/does not improve on the former best one) to reach state of the art results. 

The proposed active ranking mechanism, inspired from recent advances in the domain of 
preference elicitation \cite{Viappiani12}, is an approximation of the Bayesian expected utility of selection criterion; on the positive side, \AEUS\ is tractable in high-dimensional continuous search spaces; on the negative side, it lacks the approximate optimality guarantees of EUS. 

A first research perspective concerns the theoretical analysis of the \APRIL\ algorithm, specifically its 
convergence and robustness w.r.t. the ranking noise, and the approximation quality of the \AEUS\ criterion. In particular, the computational 
effort of \AEUS\ could be reduced with no performance loss by using Berstein races to decrease the number of empirical estimates (considered trajectories in Eq. \ref{eq:ct}) and confidently discard unpromising solutions \cite{Verena09}. 

Another research perspective is related to a more involved analysis of the expert's preferences. 
Typically, the expert might (dis)like a trajectory because of some fragments of it (as opposed to, the whole of it). Along this line, a multiple-instance ranking setting \cite{MIP} could be used to 
learn preferences at the fragment (sub-behavior) level, thus making steps toward the definition of
sub-behaviors and modular RL. 

Another further work will be concerned with hybrid policies, combining the (NN-based) parametric policy
and the model of the expert's preferences. The idea behind such hybrid policies is to provide the 
agent with both reactive and deliberative skills: while the action selection is achieved by the parametric policy by default, the expert's preferences might be exploited to reconsider these actions in some (discretized) sensori-motor states. 

On the applicative side, \APRIL\ will be experimented on large-scale robotic problems, where designing good reward functions is notoriously difficult. 

\\

\textbf{Acknowledgments}.
The first author is funded by FP7 European Project {\em Symbrion}, FET IP 216342, \url{http://symbrion.eu/}. This work is also partly funded by ANR Franco-Japanese project Sydinmalas ANR-08-BLAN-0178-01.

\bibliographystyle{splncs03}
\bibliography{ourbib}
\end{document}